\documentclass[10pt,twocolumn,letterpaper]{article}

\usepackage{iccv}
\usepackage{times}
\usepackage{epsfig}
\usepackage{graphicx}
\usepackage{amsmath}
\usepackage{amssymb}
\usepackage{subfigure}
\usepackage{booktabs}
\usepackage{caption}

\usepackage[breaklinks=true,bookmarks=false]{hyperref}

\iccvfinalcopy 

\def\iccvPaperID{****} 

\ificcvfinal\fi

\begin{document}

\title{TPPI-Net:  Towards Efficient and Practical Hyperspectral Image Classification}

\author{Hao Chen$^1$ \qquad Xiaohua Li$^1$ \qquad Jiliu Zhou$^1$ \\
  $^1$College of Computer Science, Sichuan University \\
  \small \url{chenhao1@stu.scu.edu.cn} \qquad \small \url{lxhw@scu.edu.cn}
}

\maketitle
\ificcvfinal\thispagestyle{empty}\fi

\begin{abstract}
   Hyperspectral Image(HSI) classification is the most vibrant field of research in the hyperspectral community, which aims to assign each pixel in the image to one certain category based on its spectral-spatial characteristics. Recently, some spectral-spatial-feature based DCNNs have been proposed and demonstrated remarkable classification performance. When facing a real HSI, however, these Networks have to deal with the pixels in the image one by one. The pixel-wise processing strategy is inefficient since there are numerous repeated calculations between adjacent pixels. In this paper, firstly, a brand new Network design mechanism TPPI (training based on pixel and prediction based on image) is proposed for HSI classification, which makes it possible to provide efficient and practical HSI classification with the restrictive conditions attached to the hyperspectral dataset. And then, according to the TPPI mechanism, TPPI-Net is derived based on the state of the art networks for HSI classification. Experimental results show that the proposed TPPI-Net can not only obtain high classification accuracy equivalent to the state of the art networks for HSI classification, but also greatly reduce the computational complexity of hyperspectral image prediction.
\end{abstract}

\section{Introduction}

Hyperspectral images consist of hundreds of narrow contiguous wavelength bands carrying a wealth of spectral information. Taking advantage of the rich spectral information, hyperspectral data are extremely useful in a wide range of applications in remote sensing, such as urban monitoring \cite{4686022}, agriculture \cite{4779060}, change or target detection \cite{4674613, 1704970}. Hyperspectral image (HSI) classification (i.e., assigning specific pixel to one certain category based on its characteristics) is the most vibrant field of research in the hyperspectral community and has drawn broad attention in the remote sensing field \cite{8697135}.

HSI classification methods can be divided into spectral-feature based methods and spectral-spatial-feature based methods according to the input information used. In early research attempts, the spectral vector of the pixel was intuitively used for classification to take advantage of abundant spectral bands. Spectral-feature based classification methods mainly concentrate on two steps: feature engineering and classification. Feature engineering methods include feature selection and feature extraction \cite{6450025}. The former identifies the best subset of spectral bands from all the candidate bands based on an adapted selection criterion \cite{1420295, 990133}. The latter generates a small number of new features using a transformation matrix based on some criterion for the optimum subspace \cite{6084704, 5942156}. Jia and Richards \cite{295042} proposed a simplified maximum likelihood classification technique for hyperspectral data. Murat Dundar and Landgrebe \cite{1176160} proposed a model-based mixture classifier, which uses mixture models to characterize class densities.  Bazi and Melgani \cite{1717732} presented an optimal SVM classification system for HSI classification. J.ham \etal \cite{1396322} presented the random forest framework for classification of HSI. Li \etal \cite{5559437} used multinomial logistic regression with active Learning for HSI processing. Villa \etal \cite{5594853} proposed using independent component discriminant analysis for HSI classification.

With the development of imaging technology, hyperspectral sensors can provide good spatial resolution. As a result, detailed spatial information has become available. It has been found that spectral-spatial based methods can provide good improvement in terms of classification performance \cite{8101519}. More and more spectral-spatial features based classification frameworks have been developed, which incorporate the spatial contextual information into pixel-wise classifiers. Benediktsson \etal \cite{1396321} proposed a method based on mathematical morphology for preprocessing the hyperspectral data. Camps-Valls \etal \cite{1576697} presented a framework of composite kernel machines for enhanced classification of hyperspectral images.

Very recently, some spectral–spatial-feature based DCNN (deep learning based convolutional neural networks) have been proposed and demonstrated remarkable classification performance. The joint deep spectral-spatial features were directly extracted by the original data via 2-D CNN for classification in \cite{8340197,8304691,8283837}.  Hamida \etal \cite{8344565} introduced 3-D CNN for HSI classification. Zhao \etal \cite{8519286} takes spectral-spatial patch-pair as input to train network, which is somewhat similar to the pixel-pair DCNN \cite{7736139}. The accuracy of the state of the art spectral-spatial based DCNNs: SSRN (spectral-spatial residual network \cite{8061020}) and pResNet (deep pyramidal residual networks \cite{8445697}) can reach to 98\% on common HSI classification dataset: IP (Indian pines), PU (University of Pavia) and SV (Salinas Valley).

Limited by the available HSI dataset, researchers mainly focused on the classification of individual hyperspectral pixels rather than the whole semantic segmentation of real hyperspectral images.

However, we are usually faced with a real hyperspectral image rather than individual hyperspectral pixel in practical applications. To assign every pixel in a HSI to one certain category, prior approaches usually adopted the way of pixel-wise processing in which each pixel is classified according to the information of its enclosing object or region. These methods are inefficient due to repeated computation of adjacent pixels. In fact, semantic segmentation networks such as FCN (Fully Convolutional Networks) \cite{Long_2015_CVPR}, which are generally applied in the nature image dense prediction, maybe more suitable for the HSI classification. 

Inspired by FCN, we propose a network construction strategy that can take into account both training and prediction: training based on pixel and prediction based on image (TPPI). Following the TPPI strategy and a few simple rules, not only the new TPPI-Net can be designed easily, but also the current spectral-spatial-feature based classification network can be easily modified to obtain the corresponding TPPI-Net. 

Training based on pixel makes TPPI-Net compatible with limited availability hyperspectral dataset, while prediction based on image can avoid redundant calculations between adjacent pixels when processing a hyperspectral image of real scene. We conducted experiments on common data sets, the experimental results show that TPPI-Net can not only obtain high classification accuracy equivalent to the state of the art HSI classification DCNN, but also greatly reduce the amount of computation when predicting the entire hyperspectral image.

To summarize, our main contributions are:\begin{itemize}
\item A mechanism named TPPI (Training based on pixel and prediction based on the image) is proposed for efficient and practical HSI classification. To our knowledge, it is the first attempt to wriggle out from under restrictive conditions attached to hyperspectral dataset and pay attention to the efficient dense prediction of a hyperspectral image. 
\item Following the TPPI mechanism, TPPI-Net is designed which is the first image to image network for HSI classification. TPPI-Net can be regarded as a classification network in the train phase and a semantic segmentation network in the prediction phase.
\item Some basic rules which should be obeyed during TPPI-Net designing are given. And rigorous tests, as well as detailed analysis, are done to fully demonstrate the effectiveness of the proposed method.
\end{itemize}

\section{Problem Analysis}

In general, a learning-based HSI classification method includes three phases: training, testing and prediction. In the training phase, a training dataset consisting of labeled pixels is used to learn the mapping model that transforms spatial-spectral information of pixel to the land cover class of the pixel. In the testing phase, a test dataset which doesn't intersect with the training dataset is used to test the model. And in the prediction phase, a real HSI is predicted by feeding the pixels in HSI into the model one by one. It's obvious that, when facing a real HSI, the way of pixel-wise processing is low efficiency since there are lots of repetitive computations between adjacent pixels. In fact, if the dataset for HSI classification is matched with the requirement for training samples, the semantic segmentation network that has received a lot of attention recently is more suitable for the HSI classification task, because it is a kind of image to image architecture instead of the image to scalar architecture used in current HSI classification.

\subsection{Semantic segmentation networks}
Semantic segmentation is the task of clustering the pixels in an image together if they belong to the same land cover type. It is a form of pixel-level prediction because it assigns each pixel in an image to one certain category. Although semantic segmentation has been an important branch in computer vision, there's virtually no great breakthrough until Long \etal first used FCN to achieve end-to-end natural images segmentation in 2014. Prior to FCN, the usual approach for dense prediction of a natural image was to use a classification network to predict pixel by pixel, which is the same idea as the current HSI dense prediction. After that, researchers have proposed various segmentation networks, such as U-Net \cite{10.1007/978-3-319-24574-4_28}, SegNet \cite{7803544}, FC-Densenet \cite{Jegou_2017_CVPR_Workshops}, E-Net \cite{paszke2016enet} and Link-Net \cite{8305148}, RefineNet \cite{Lin_2017_CVPR}, PSPNet \cite{Zhao_2017_CVPR}, Mask-RCNN \cite{He_2017_ICCV}, and some semi-supervised methods, such as DecoupledNet \cite{hong2015decoupled}.
Figure 1 shows the architecture of SegNet. As can be seen from Figure 1, the semantic segmentation network is an image to image architecture. That is, if the train sample is an image with $w*h$ size, its classification map should also be an image with the same size.
\begin{figure}[h]
	\begin{center}
   		\includegraphics[width=1\linewidth]{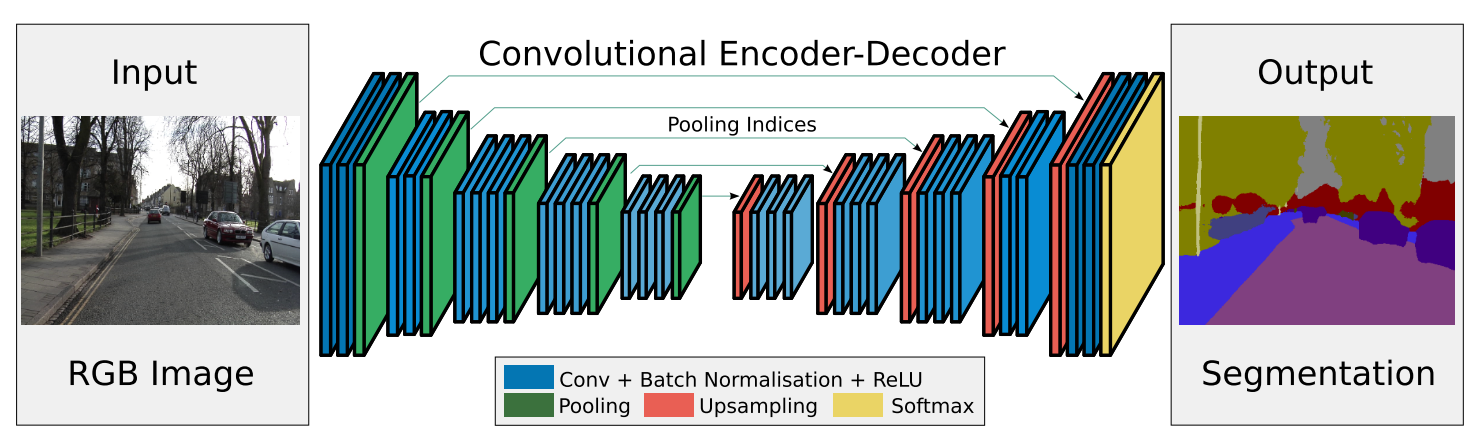}
	\end{center}
   \caption{The structure of SegNet.}
\end{figure}
\subsection{The limitation of General Dataset for HSI classification}
For HSI classification, the common datasets only have one hyperspectral image and a corresponding GT in which only some pixels (no more than 50\%) are labeled.Therefore, a dataset for HSI classification can be regarded as the sets of labeled pixels, rather than the sets of pixel-wise labeled patches. 
Consider a hyperspectral data set with M labeled pixels, it is denoted as following:
\begin{equation}
    X=\{x_i,y_i \}_{i=1}^{M}
\end{equation}
Where $x_i\in{Z^{m*m*B}}$ is the $i^{th}$ sample, and  $y_i\in\{1,2,...,C\}$ is the label of $x_i$ , $B$ represents the number of bands and $C$ represents the number of classes, m is the spatial size of neighborhood.

It is clear that the label $y$ of a sample $x$ in the dataset is a scalar, rather than an image with the same spatial size as the sample $x$.

The mismatching between the image-image requirement of the semantic segmentation networks and the image-scalar characteristic of the HSI dataset hold back the direct employment of the existing semantic segmentation networks for the efficient HSI classification.

In order to realize the efficiency HSI classification in the prediction phase, and take the nature of the dataset into account at the same time, a mechanism named TPPI (Training based on pixel and prediction based on the image) is proposed in the next section. And then the corresponding TPPI-Net is designed, which can be regarded as a classification network in the training phase and a semantic segmentation network in the prediction phase.


\section{Methods}
\subsection{The framework of proposed HSI classification based on TPPI-Net}
\begin{figure*}
\centering
\subfigure[TPPI]{
\begin{minipage}[t]{0.5\linewidth}
\centering
\includegraphics[width=0.8\linewidth]{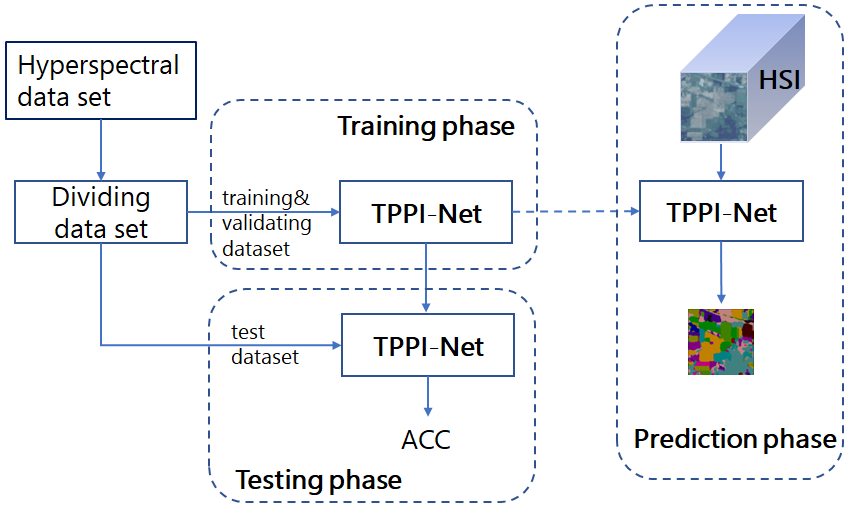}
\end{minipage}%
}%
\subfigure[TPPP]{
\begin{minipage}[t]{0.5\linewidth}
\centering
\includegraphics[width=0.8\linewidth]{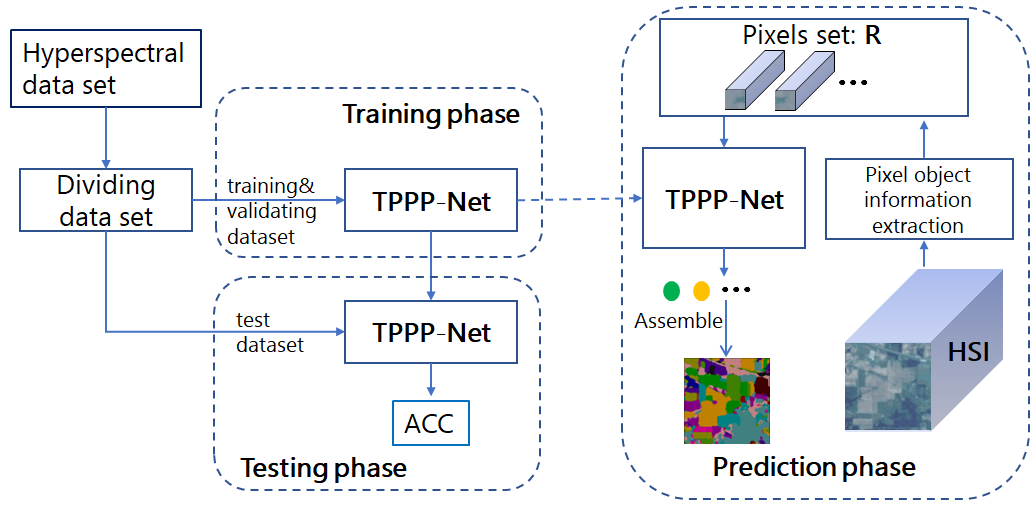}
\end{minipage}%
}%
\centering
\caption{Flowchart of : (a). TPPI mechanism and  (b). TPPP mechanism.}
\end{figure*}
In order to realize the efficiency HSI classification in the prediction phase, and take the nature of the dataset into account at the same time, this paper proposed a novel mechanism named TPPI for HSI classification.
 
Figure 2(a) is the flowchart of the HSI classification method based on the TPPI. And Figure 2(b) shows the process of the existing DCNN based HSI method, which is called TPPP mechanism (Training based on Pixels and Prediction based on Pixels) here. 

As shown in Figure 2, during the training and testing phase, the processing procedures of the two mechanisms are exactly the same. That is, the input is a sample with spatial size m*m and the output is a scalar. 

However, in the prediction phase, that is, the actual application phase, the two methods are totally different. The input of TPPI-Net could be a hyperspectral image with any size, and the output of TPPI-Net is the classification map of original HSI. While the input of TPPP-Net must be a sample with spatial size m*m and the output of TPPP-Net is still a scalar. 
  
TPPP-Net mainly focused on the classification of individual hyperspectral pixels and ignores the computational complexity in the prediction phase.  
TPPI-Net  proposed in this paper not only takes into account the limitation of general dataset in the training and testing phase, but also pursues the efficient HSI classification in the practical applications.
\subsection{Architecture design of TPPI-Net}
\begin{figure*}
\centering
\subfigure[SSRN]{
\begin{minipage}[t]{0.8\linewidth}
\centering
\includegraphics[width=0.8\linewidth]{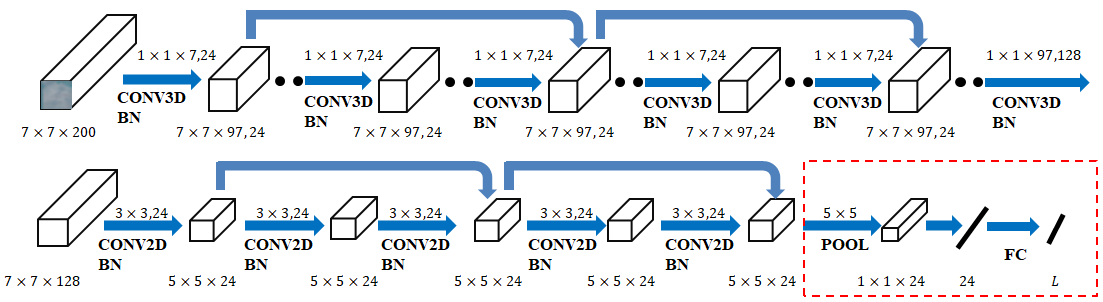}
\end{minipage}%
}%

\subfigure[SSRN-TPPI]{
\begin{minipage}[t]{0.8\linewidth}
\centering
\includegraphics[width=0.8\linewidth]{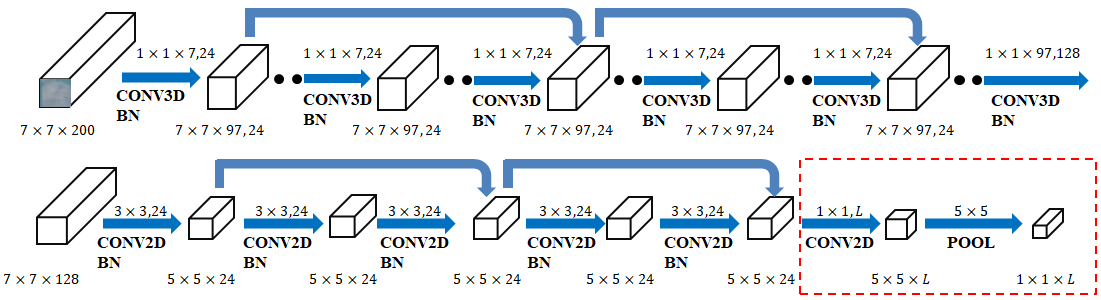}
\end{minipage}%
}%
\centering
\caption{Graphical illustration of SSRN and SSRN-TPPI.}
\end{figure*}

\begin{figure*}
\centering
\subfigure[pResNet]{
\begin{minipage}[t]{0.5\linewidth}
\centering
\includegraphics[width=0.8\linewidth]{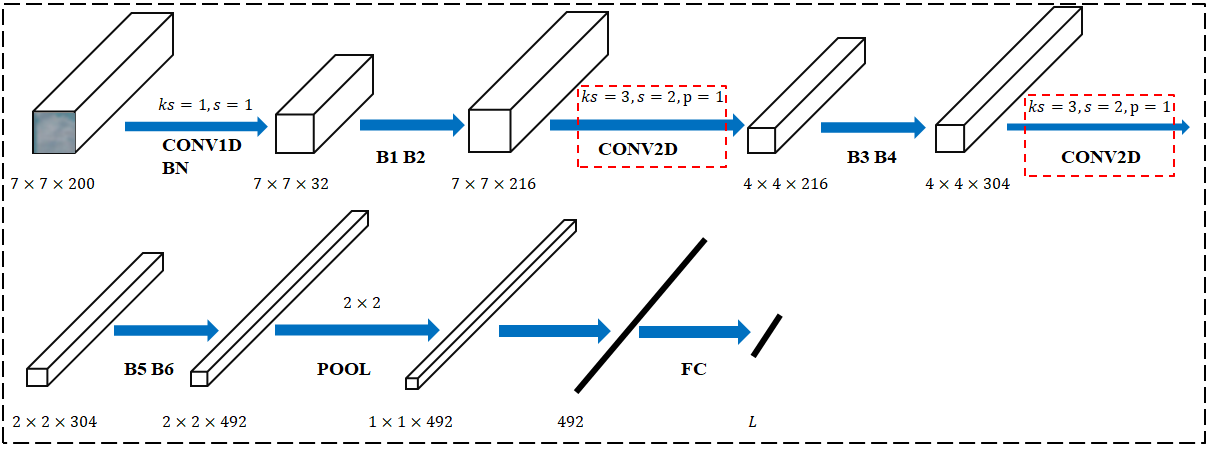}
\end{minipage}%
}%
\subfigure[pResNet-TPPI]{
\begin{minipage}[t]{0.5\linewidth}
\centering
\includegraphics[width=0.8\linewidth]{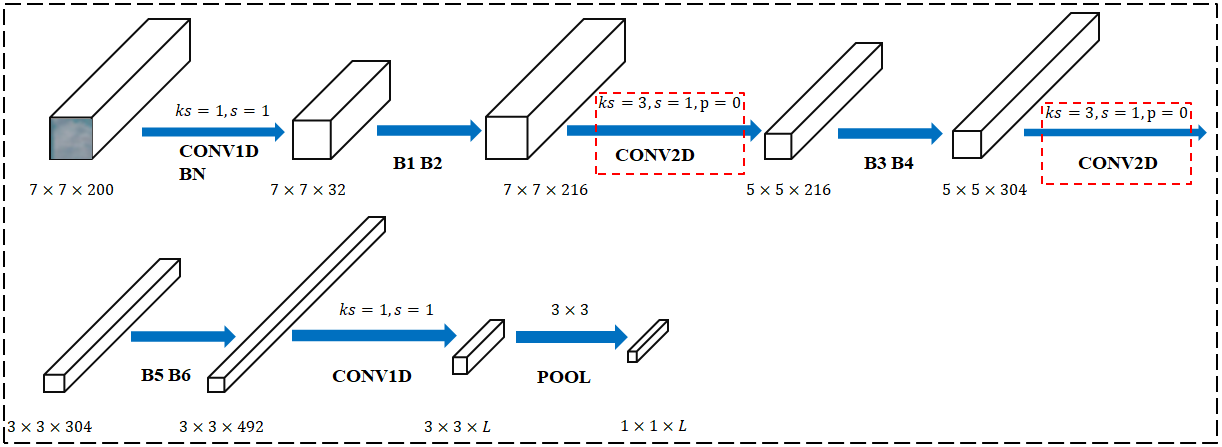}
\end{minipage}%
}%

\subfigure[BottleNeck]{
\begin{minipage}[t]{0.8\linewidth}
\centering
\includegraphics[width=0.5\linewidth]{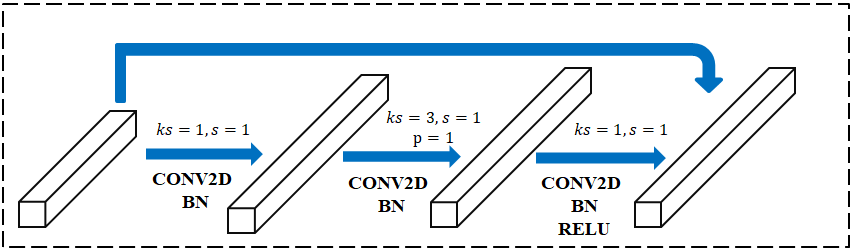}
\end{minipage}%
}%

\centering
\caption{(a) and (b) is the graphical illustration of pResNet and pResNet-TPPI. (c) is the structure of bottleneck: the B1 $\sim$ B6 in (a) and (b).}
\end{figure*}

\begin{figure}
\centering
\subfigure[]{
\begin{minipage}[t]{0.3\linewidth}
\centering
\includegraphics[width=0.8\linewidth]{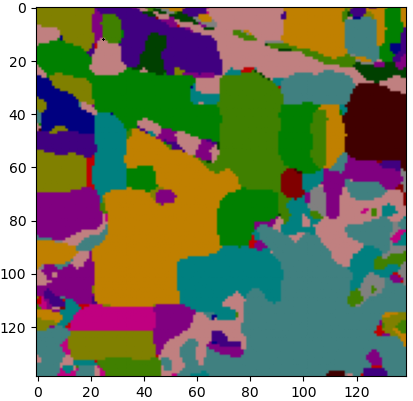}
\end{minipage}%
}%
\subfigure[]{
\begin{minipage}[t]{0.3\linewidth}
\centering
\includegraphics[width=0.8\linewidth]{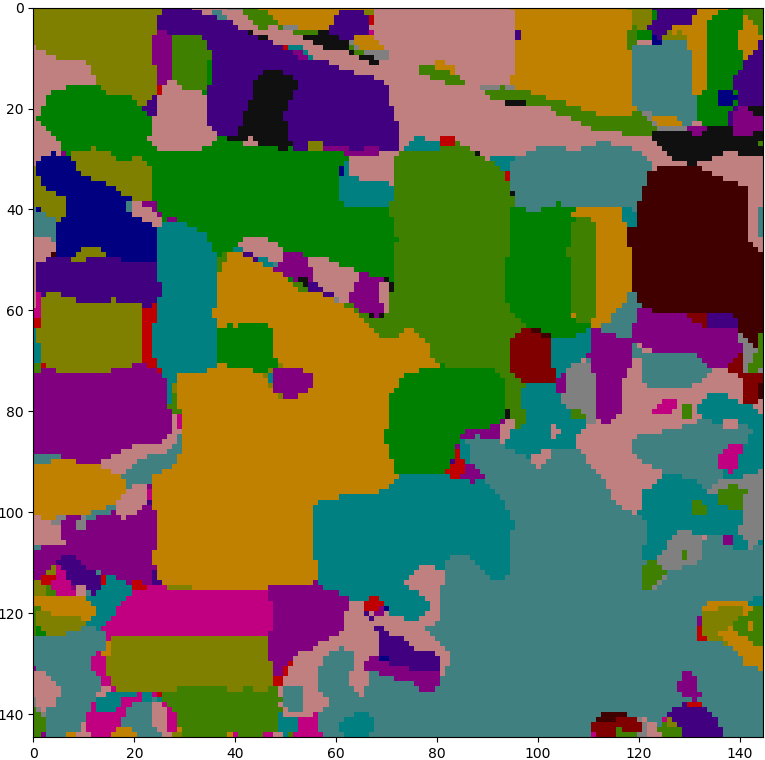}
\end{minipage}%
}%
\subfigure[]{
\begin{minipage}[t]{0.3\linewidth}
\centering
\includegraphics[width=0.8\linewidth]{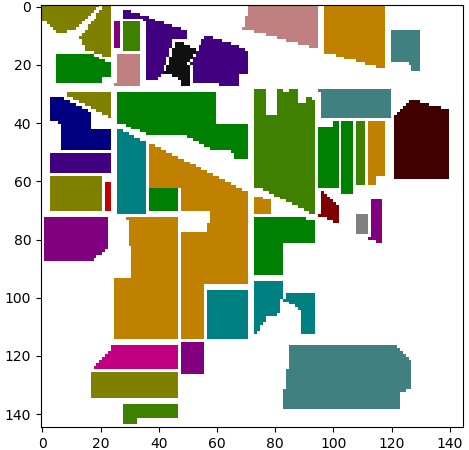}
\end{minipage}%
}%
\centering
\caption{(a)classification map of IP (size: 139*139). (b) classification map of IP (size: 145*145). (c) Ground Truth (size: 145*145).}
\end{figure}

As we know, a scalar can be seen as a special form of an image with size $1*1$. So, the hyperspectral dataset with M labeled pixels formulated in $Eq. (1)$ can be rewritten as following:
\begin{equation}
   X=\{x_i,y_i \}_{i=1}^{M}
\end{equation}
Where $x_i\in{Z^{m*m*B}}$ is the $i^{th}$ sample, and  $y_i\in{L^{(m-m+1)*(m-m+1)}}$ is the label of $x_i$. $L(=\{1,2,…C\})$ is a set. $B$ represents the number of bands, $C$ represents the number of classes, $m$ is the spatial size of neighborhood of the labeled pixel.

Based on the presentation above, we can deduce that, for a TPPI-Net, given the spatial size of the samples in the dataset is $m*m$, if the spatial size of input is $W*H$, then the size of the output should be $(W-m+1)*(W-m+1)$ whether in training, testing or prediction phase.

As mentioned above, for a TPPI-Net, both the input and output can be regarded as image and they have a constant relationship in terms of spatial size.  In order to meet the constraint, the following rules should be obeyed:

(1). No FC (fully connected) layer in whole Net, since FC layer will make image-to-image network impossible.

(2). No down-sample in whole Net. Because limited availability and the small spatial size of training samples for HSI classification will not only restrict the depth of the Net, but also make encode-decode architecture shown in Figure 1 unrealistic.  The size reduction from input to output should be done by setting the parameter $padding = 0$ in some layers.

Following the above rules, not only the new TPPI-Net can be designed easily, but also the current spectral-spatial-feature based classification network can be easily modified to obtain the corresponding TPPI-Net.

In order to facilitate subsequent comparative experiments, we modified two state of the art HSI classification networks (pResNet and SSRN) and obtain two TPPI-Nets (SSRN-TPPI and pResNet-TPPI).

Figure 3 shows the structure of SSRN and SSRN-TPPI. In order to get SSRN-TPPI, we replaced the fully connected layer in SSRN using a convolutional layer. Figure 4 shows the structure of pResNet and pResNet-TPPI. Not only is the fully connected layer in pResNet substituted by a convolutional layer, but also all the downsampling layer in pResNet are removed.  Specifically, all the conv2d layers with $stride=2$ are replaced by conv2d layers with $padding=0$ and $stride=1$. 

In fact, following the rules mentioned above, we can modify any spectral-spatial-feature based HSI classification networks to obtain the corresponding TPPI-Net. 

Taking Figure 3 as an example, the training, testing and prediction phase of TPPP-Net and TPPI-Net are demonstrated in detail.

In the training and testing phase, the dataset used by SSRN is pixel-based as formulated in $Eq.  (1)$. The input of SSRN is a patch $(7*7*200)$ and the output is a scalar which denotes the land cover class of the central pixels of the input patch.

The attendance of FC layer makes that the size of input must be identical in the prediction phase with that in the training phase.

In the training and testing phase, the dataset used by SSRN-TPPI is also pixel-based as formulated in $Eq. (2)$, the input of SSRN-TPPI is also a patch $(7*7*200)$ and the output is a scalar which indeed can be regarded as a smaller patch $((7-7+1)*(7-7+1))$.

In the prediction phase, the input of SSRN-TPPI can be any size because it is a full convolution network. There is a fixed relationship between output size $(W^{'}*H^{'})$ and input size$(W*H)$: $W^{'}=W-m+1$, $H^{'}=H-m+1$. $m$ is the spatial size of neighborhood of the labeled pixel which is 7 in Figure 3.

Figure 5(a) shows the classification map $(139*139)$ of IP image $(145*145)$ predicted by SSRN-TPPI $(m=7)$. In order to get the original size $(145*145)$ classification map, one can expand the original HSI by padding to $(151*151)$ before feeding it to the Net. The corresponding classification map is shown in Figure 5(b).
\subsection{Complexity analysis of proposed method}
Compared with the HSI classification method based on TPPP-Net, the HSI classification method based on TPPI-Net can avoid repeated calculations of adjacent pixels in the prediction phase and obtain a faster speed.

Still, we take SSRN and SSRN-TPPI as examples to analyze the computational complexity.

Although the two networks in Figure 3 seemed similar, when predicting a real HSI, FLOPs of SSRN-TPPI is much smaller than that of SSRN. Computation complexity of the two Nets is analyzed by taking a conv3d layer and a conv2d layer as examples. Where $H*W$ is the size of the real HSI which will be dense predicted, $7*7$ is the size of sample during the training phase.

1).Take the $2nd$ layer ($1*1*7$ 3D convolutional layer) in SSRN-TPPI and SSRN in Figure 3 as an example: To predict the entire image, the FLOPs of SSRN-TPPI is: $H*W*97*24*7*24$.To predict one pixel, the FLOPs of SSRN is: $7*7*97*24*7*24$. To predict the entire image, the FLOPs of SSRN is: $H*W*7*7*97*24*7*24$. It can be seen that the FLOPs of SSRN is $7*7$ times that of SSRN-TPPI.

2).Take the $7th$ layer ($3*3$ 2D convolutional layer) in SSRN-TPPI and SSRN in Figure 3 as an example: To predict the entire image, the FLPOs of SSRN-TPPI is: $H*W*128*3*3*24$. To predict one pixel, the FLOPs SSRN is: $7*7*128*3*3*24$. Then, to predict the entire image, the FLOPs SSRN is: $H*W*7*7*128*3*3*24$. It can be seen that the FLOPs of SSRN is $7*7$ times that of SSRN-TPPI.

In addition, from the analysis above, we can note, given a real HSI, with the increase of train patch size, the FLOPs of SSRN-TPPI will remain unchanged, while the FLOPs of SSRN will grow dramatically $(O(m2))$.

According to the above theoretical analysis, TPPI-Net will greatly improve the speed of HSI dense prediction, compared with TPPP-Net.

\section{Experiments and results}
\subsection{Dataset}
In our study, three well-known hyperspectral datasets with different environmental settings were adopted to validate the proposed TPPI-Net. They are a mixed vegetation site over the Indian Pines test area in Northwestern Indiana (Indian Pines), an urban site over the city of Pavia, Italy (University of Pavia), and a site over Salinas Valley (SV), California.

\textbf{IP} : This scene was gathered by AVIRIS sensor over the Indian Pines test site in North-western Indiana and consists of $145*145$ pixels and 224 spectral reflectance bands in the wavelength range 400–2500 nanometers. Although the image contains 21025 pixels, of which only 10249 pixels are labeled into sixteen classes in the ground truth. The number of bands is reduced to 200 by removing bands covering the region of water absorption.

\textbf{PU} : This scene was acquired by the ROSIS sensor during a flight campaign over Pavia, northern Italy. The number of spectral bands is 103 for Pavia University. Pavia University concludes a $610*340$ image, and only 42776 pixels of the total 207400 pixels are labeled into nine classes in the ground truth.

\textbf{SV} : This scene was collected by the 224-bands AVIRIS sensor over Salinas Valley, California, and is characterized by high spatial resolution (3.7meter pixels). As with Indian Pines scene, the bands of Salinas Valley is reduce to 203 after removing bands covering the region of water absorption. Salinas’ ground truth contains 16 classes. This data set contains 111104 pixels and only 54129 pixels are labeled.
\subsection{Experiment Configuration}
We need to select some labeled pixels on the HSI as our training dataset because each common HSI data set contains only one image. Taking the IP data set as an example, the specific data set division method is as follows: Unlabeled points are not used for each class, 20\% and 16\% of the labeled pixels are selected at random as the training set and validation set respectively, and the remaining as the test set. 

In order to verify the performance of TPPI-Nets, we designed experiments to compare the classification accuracy and prediction time of SSRN-TPPI and SSRN, pResNet and pResNet-TPPI.

Overall accuracy (OA), average accuracy (AA), and kappa coefficient (Kappa) are used to evaluate the classification accuracy. For TPPP-Nets : SSRN and pResNet, which employ the way of pixel-wise processing when predicting an image, we directly calculate their OA AA Kappa on the test dataset. For TPPI-Nets : SSRN-TPPI and pResNet-TPPI, the classification accuracy is calculated not only in the test phase but also in the prediction phase, because receptive field in prediction phase is a little different with that in test phase. It is worth noting that the unlabeled pixels and pixels used for training and verification will not be included when calculating the classification accuracy in the prediction phase.

Running time on GPU and CPU for  an entire HSI (including unlabeled points, training and validation dataset pixels) are used to evaluate the prediction speed. 

During the train phase, the size of mini-batch was 100, the learning rate was 0.01, the momentum was 0.9 and the weight decay was 0.0001, the number of training epochs is 200. All experiments are conducted on a computer with the GTX 1080 Ti graphical processing unit (GPU), what’s more, we also conducted an additional prediction time experiment on the CPU.
\subsection{Experimental Results and Discussion}
\begin{table}
  \centering
  \caption{The classification accuracy of SSRN, SSRN-TPPI, pResNet and pResNet-TPPI on the IP dataset in the test and prediction phase.}
\resizebox{0.8\linewidth}{18mm}{
    \begin{tabular}{c|c|cc|c|cc}
    \toprule
    \toprule
          & \multicolumn{1}{p{5.5em}|}{SSRN} & \multicolumn{2}{p{11em}|}{SSRN-TPPI} & \multicolumn{1}{p{5.5em}|}{pResNet} & \multicolumn{2}{p{8.38em}}{pResNet-TPPI} \\
\cmidrule{2-7}          & \multicolumn{1}{p{5.5em}|}{test \& prediction} & \multicolumn{1}{p{5.5em}}{test} & \multicolumn{1}{p{5.5em}|}{prediction} & \multicolumn{1}{p{5.5em}|}{test \& prediction} & \multicolumn{1}{p{4.19em}}{test} & \multicolumn{1}{p{4.19em}}{prediction} \\
    \midrule
    \multicolumn{1}{p{4.19em}|}{\textbf{OA(\%)}} & \multicolumn{1}{p{5.5em}|}{\textbf{98.58±0.06}} & \multicolumn{1}{p{5.5em}}{\textbf{98.60±0.05}} & \multicolumn{1}{p{5.5em}|}{\textbf{99.19±0.13}} & \multicolumn{1}{p{5.5em}|}{\textbf{99.11±0.07}} & \multicolumn{1}{p{4.19em}}{\textbf{99.24±0.12}} & \multicolumn{1}{p{4.19em}}{\textbf{98.73±0.03}} \\
    \multicolumn{1}{p{4.19em}|}{\textbf{AA(\%)}} & \multicolumn{1}{p{5.5em}|}{\textbf{98.15±0.15}} & \multicolumn{1}{p{5.5em}}{\textbf{97.93±0.11}} & \multicolumn{1}{p{5.5em}|}{\textbf{98.63±0.33}} & \multicolumn{1}{p{5.5em}|}{\textbf{98.38±0.22}} & \multicolumn{1}{p{4.19em}}{\textbf{97.05±1.43}} & \multicolumn{1}{p{4.19em}}{\textbf{98.10±0.04}} \\
    \multicolumn{1}{p{4.19em}|}{\textbf{Kappa}} & \multicolumn{1}{p{5.5em}|}{\textbf{98.39±0.07}} & \multicolumn{1}{p{5.5em}}{\textbf{98.40±0.05}} & \multicolumn{1}{p{5.5em}|}{\textbf{99.08±0.15}} & \multicolumn{1}{p{5.5em}|}{\textbf{98.99±0.09}} & \multicolumn{1}{p{4.19em}}{\textbf{99.13±0.14}} & \multicolumn{1}{p{4.19em}}{\textbf{98.55±0.04}} \\
    \midrule
    \textbf{1} & 96.43 & 96.43 & 97.14 & 92.86 & 90    & 96.43 \\
    \textbf{2} & 98.32 & 98.54 & 99.32 & 99.63 & 99.58 & 98.1 \\
    \textbf{3} & 98.55 & 99    & 98.79 & 98.46 & 98.27 & 99.69 \\
    \textbf{4} & 89.84 & 88.08 & 92.18 & 91.52 & 94.84 & 88.74 \\
    \textbf{5} & 100   & 100   & 99.94 & 99.68 & 99.74 & 100 \\
    \textbf{6} & 99.79 & 99.79 & 99.57 & 99.44 & 99.36 & 99.86 \\
    \textbf{7} & 96.08 & 96.08 & 98.82 & 96.47 & 94.12 & 94.12 \\
    \textbf{8} & 100   & 100   & 100   & 100   & 100   & 100 \\
    \textbf{9} & 100   & 97.22 & 96.67 & 100   & 81.67 & 100 \\
    \textbf{10} & 98.71 & 98.6  & 98.68 & 98.49 & 98.71 & 98.6 \\
    \textbf{11} & 98.7  & 98.6  & 99.26 & 99.27 & 99.59 & 98.56 \\
    \textbf{12} & 95.25 & 95.95 & 99.37 & 98.52 & 99.1  & 98.16 \\
    \textbf{13} & 100   & 100   & 100   & 99.85 & 99.85 & 100 \\
    \textbf{14} & 99.59 & 99.55 & 99.9  & 99.95 & 99.98 & 99.47 \\
    \textbf{15} & 99.19 & 99.05 & 99.51 & 99.84 & 100   & 99.59 \\
    \textbf{16} & 100   & 100   & 98.99 & 100   & 97.97 & 98.31 \\
    \bottomrule
    \bottomrule
    \end{tabular}}%
  \label{tab:addlabel}%
\end{table}%

\begin{table}
  \centering
  \caption{The classification accuracy of SSRN, SSRN-TPPI, pResNet and pResNet-TPPI on the PU dataset in the test and prediction phase.}
\resizebox{0.8\linewidth}{12mm}{
    \begin{tabular}{c|c|cc|c|cc}
    \toprule
    \toprule
          & \multicolumn{1}{p{5.5em}|}{SSRN} & \multicolumn{2}{p{11em}|}{SSRN-TPPI} & \multicolumn{1}{p{5.5em}|}{pResNet} & \multicolumn{2}{p{8.38em}}{pResNet-TPPI} \\
\cmidrule{2-7}          & \multicolumn{1}{p{5.5em}|}{test \& prediction} & \multicolumn{1}{p{5.5em}}{test} & \multicolumn{1}{p{5.5em}|}{prediction} & \multicolumn{1}{p{5.5em}|}{test \& prediction} & \multicolumn{1}{p{4.19em}}{test} & \multicolumn{1}{p{4.19em}}{prediction} \\
    \midrule
    \multicolumn{1}{p{4.19em}|}{\textbf{OA(\%)}} & \multicolumn{1}{p{5.5em}|}{\textbf{99.46±0.03}} & \multicolumn{1}{p{5.5em}}{\textbf{99.47±0.01}} & \multicolumn{1}{p{5.5em}|}{\textbf{99.44±0.02}} & \multicolumn{1}{p{5.5em}|}{\textbf{99.87±0.02}} & \multicolumn{1}{p{4.19em}}{\textbf{99.85±0.04}} & \multicolumn{1}{p{4.19em}}{\textbf{99.87±0.02}} \\
    \multicolumn{1}{p{4.19em}|}{\textbf{AA(\%)}} & \multicolumn{1}{p{5.5em}|}{\textbf{99.12±0.06}} & \multicolumn{1}{p{5.5em}}{\textbf{99.20±0.01}} & \multicolumn{1}{p{5.5em}|}{\textbf{99.17±0.05}} & \multicolumn{1}{p{5.5em}|}{\textbf{99.79±0.02}} & \multicolumn{1}{p{4.19em}}{\textbf{99.71±0.13}} & \multicolumn{1}{p{4.19em}}{\textbf{99.84±0.03}} \\
    \multicolumn{1}{p{4.19em}|}{\textbf{Kappa}} & \multicolumn{1}{p{5.5em}|}{\textbf{99.28±0.05}} & \multicolumn{1}{p{5.5em}}{\textbf{99.29±0.02}} & \multicolumn{1}{p{5.5em}|}{\textbf{99.26±0.03}} & \multicolumn{1}{p{5.5em}|}{\textbf{99.84±0.02}} & \multicolumn{1}{p{4.19em}}{\textbf{99.81±0.06}} & \multicolumn{1}{p{4.19em}}{\textbf{99.82±0.03}} \\
    \midrule
    \textbf{1} & 99.69 & 99.69 & 99.54 & 99.97 & 99.95 & 99.56 \\
    \textbf{2} & 99.99 & 99.98 & 99.95 & 99.98 & 99.98 & 99.99 \\
    \textbf{3} & 95.73 & 97.02 & 97.62 & 98.99 & 98.57 & 99.58 \\
    \textbf{4} & 99.37 & 99.35 & 98.04 & 99.88 & 99.95 & 99.97 \\
    \textbf{5} & 99.96 & 100   & 99.69 & 100   & 100   & 100 \\
    \textbf{6} & 99.65 & 99.63 & 100   & 99.96 & 99.97 & 100 \\
    \textbf{7} & 99.92 & 99.92 & 100   & 99.93 & 99.39 & 99.98 \\
    \textbf{8} & 97.79 & 97.24 & 97.72 & 99.47 & 99.62 & 99.52 \\
    \textbf{9} & 99.94 & 99.94 & 100   & 100   & 99.93 & 99.97 \\
    \bottomrule
    \bottomrule
    \end{tabular}}%
  \label{tab:addlabel}%
\end{table}%

\begin{table}
  \centering
  \caption{The classification accuracy of SSRN, SSRN-TPPI, pResNet and pResNet-TPPI on the SV dataset in the test and prediction phase.}
\resizebox{0.8\linewidth}{18mm}{
    \begin{tabular}{c|c|cc|c|cc}
    \toprule
    \toprule
          & \multicolumn{1}{p{5.5em}|}{SSRN} & \multicolumn{2}{p{11em}|}{SSRN-TPPI} & \multicolumn{1}{p{5.5em}|}{pResNet} & \multicolumn{2}{p{8.38em}}{pResNet-TPPI} \\
\cmidrule{2-7}          & \multicolumn{1}{p{5.5em}|}{test \& prediction} & \multicolumn{1}{p{5.5em}}{test} & \multicolumn{1}{p{5.5em}|}{prediction} & \multicolumn{1}{p{5.5em}|}{test \& prediction} & \multicolumn{1}{p{4.19em}}{test} & \multicolumn{1}{p{4.19em}}{prediction} \\
    \midrule
    \multicolumn{1}{p{4.19em}|}{\textbf{OA(\%)}} & \multicolumn{1}{p{5.5em}|}{\textbf{98.34±0.06}} & \multicolumn{1}{p{5.5em}}{\textbf{98.33±0.06}} & \multicolumn{1}{p{5.5em}|}{\textbf{98.84±0.40}} & \multicolumn{1}{p{5.5em}|}{\textbf{99.89±0.03}} & \multicolumn{1}{p{4.19em}}{\textbf{99.90±0.04}} & \multicolumn{1}{p{4.19em}}{\textbf{98.36±0.62}} \\
    \multicolumn{1}{p{4.19em}|}{\textbf{AA(\%)}} & \multicolumn{1}{p{5.5em}|}{\textbf{99.23±0.02}} & \multicolumn{1}{p{5.5em}}{\textbf{99.27±0.03}} & \multicolumn{1}{p{5.5em}|}{\textbf{99.46±0.16}} & \multicolumn{1}{p{5.5em}|}{\textbf{99.94±0.02}} & \multicolumn{1}{p{4.19em}}{\textbf{99.93±0.02}} & \multicolumn{1}{p{4.19em}}{\textbf{99.22±0.30}} \\
    \multicolumn{1}{p{4.19em}|}{\textbf{Kappa}} & \multicolumn{1}{p{5.5em}|}{\textbf{98.16±0.07}} & \multicolumn{1}{p{5.5em}}{\textbf{98.14±0.06}} & \multicolumn{1}{p{5.5em}|}{\textbf{98.71±0.45}} & \multicolumn{1}{p{5.5em}|}{\textbf{99.88±0.03}} & \multicolumn{1}{p{4.19em}}{\textbf{99.89±0.04}} & \multicolumn{1}{p{4.19em}}{\textbf{98.17±0.69}} \\
    \midrule
    \textbf{1} & 99.77 & 100   & 100   & 100   & 100   & 100 \\
    \textbf{2} & 100   & 99.97 & 99.9  & 100   & 100   & 100 \\
    \textbf{3} & 99.87 & 99.92 & 99.84 & 99.98 & 100   & 100 \\
    \textbf{4} & 99.85 & 99.85 & 98.51 & 100   & 99.98 & 99.89 \\
    \textbf{5} & 98.81 & 98.75 & 99.03 & 99.89 & 99.92 & 99.95 \\
    \textbf{6} & 100   & 99.96 & 99.96 & 100   & 100   & 100 \\
    \textbf{7} & 99.97 & 99.83 & 99.75 & 99.99 & 99.93 & 99.95 \\
    \textbf{8} & 96.49 & 95.06 & 95.17 & 99.75 & 99.75 & 99.96 \\
    \textbf{9} & 99.95 & 100   & 100   & 100   & 100   & 100 \\
    \textbf{10} & 99.87 & 99.71 & 99.95 & 100   & 99.99 & 100 \\
    \textbf{11} & 99.61 & 99.85 & 100   & 99.88 & 99.74 & 99.85 \\
    \textbf{12} & 100   & 100   & 100   & 100   & 100   & 100 \\
    \textbf{13} & 99.94 & 100   & 100   & 100   & 100   & 100 \\
    \textbf{14} & 100   & 99.9  & 99.61 & 100   & 100   & 100 \\
    \textbf{15} & 93.97 & 96.15 & 99.89 & 99.64 & 99.85 & 87.92 \\
    \textbf{16} & 99.6  & 99.42 & 99.71 & 99.91 & 99.71 & 99.95 \\
    \bottomrule
    \bottomrule
    \end{tabular}}%
  \label{tab:addlabel}%
\end{table}%

\begin{figure*}
\centering
\subfigure[SSRN]{
\begin{minipage}[t]{0.15\linewidth}
\centering
\includegraphics[width=0.8\linewidth]{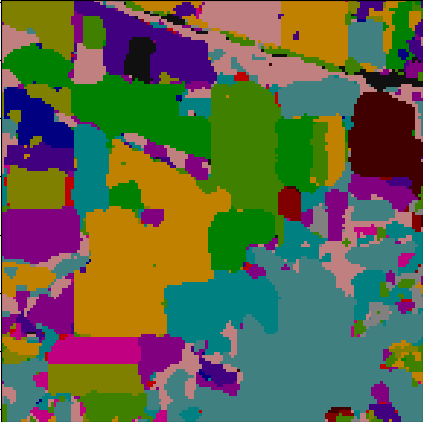}
\end{minipage}%
}%
\subfigure[SSRN-TPPI]{
\begin{minipage}[t]{0.15\linewidth}
\centering
\includegraphics[width=0.8\linewidth]{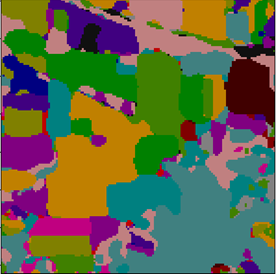}
\end{minipage}%
}%
\subfigure[pResNet]{
\begin{minipage}[t]{0.15\linewidth}
\centering
\includegraphics[width=0.8\linewidth]{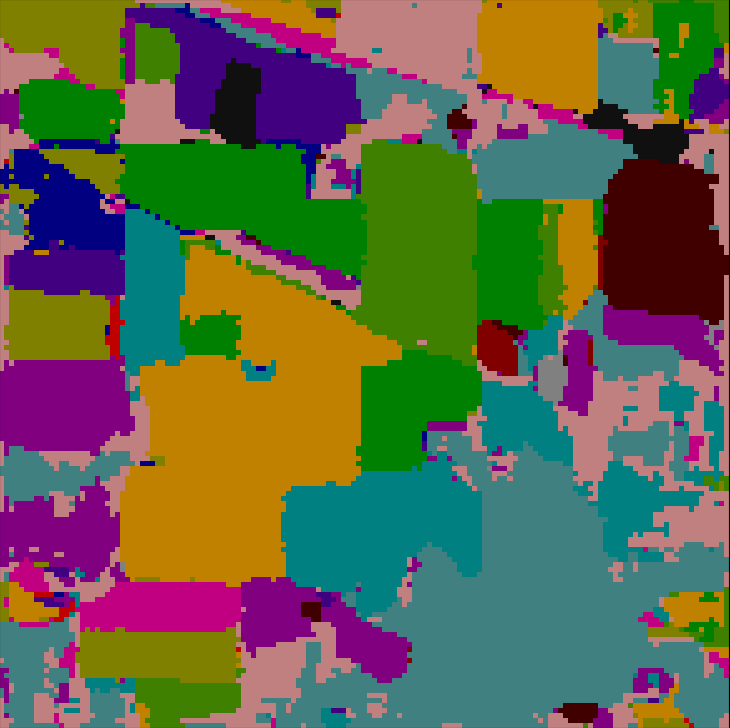}
\end{minipage}%
}%
\subfigure[pResNet-TPPI]{
\begin{minipage}[t]{0.15\linewidth}
\centering
\includegraphics[width=0.8\linewidth]{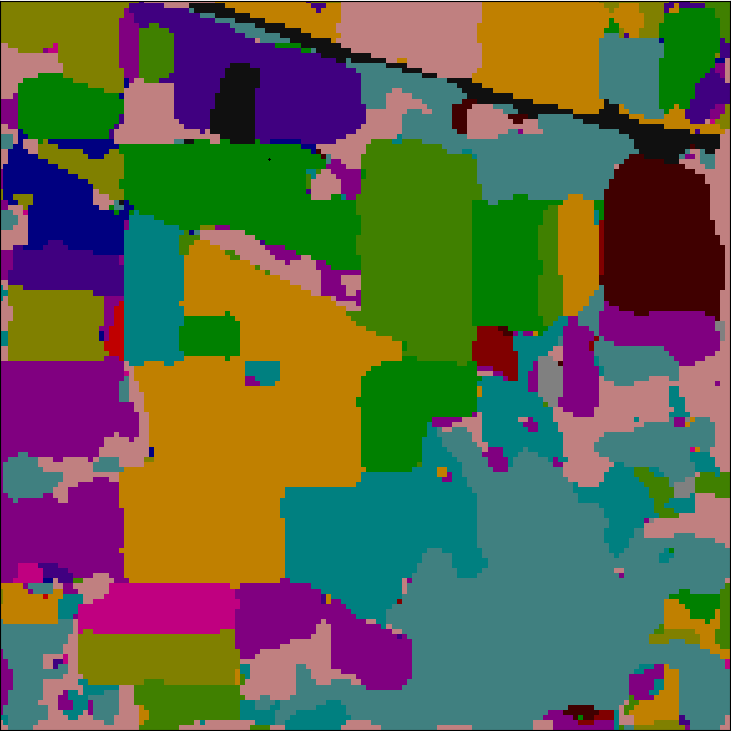}
\end{minipage}%
}%
\subfigure[GT]{
\begin{minipage}[t]{0.15\linewidth}
\centering
\includegraphics[width=0.8\linewidth]{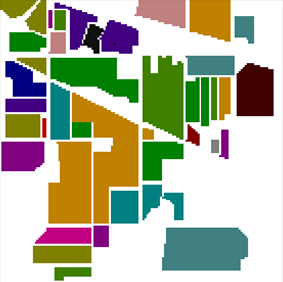}
\end{minipage}%
}%

\subfigure[SSRN]{
\begin{minipage}[t]{0.15\linewidth}
\centering
\includegraphics[width=0.8\linewidth]{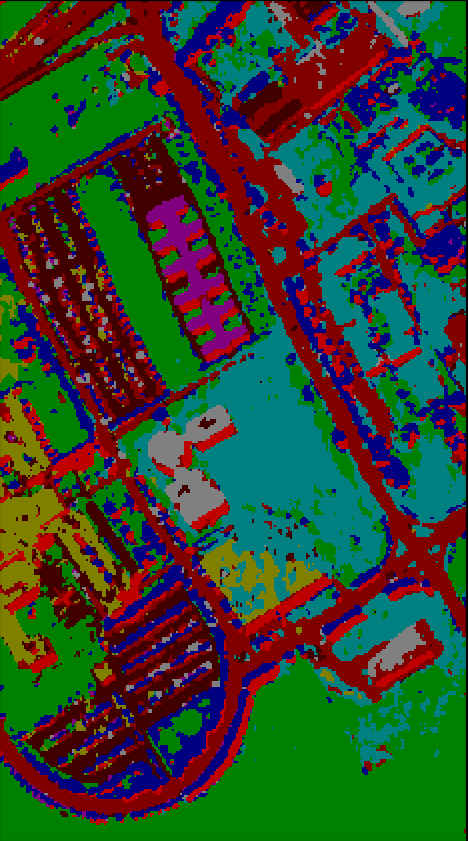}
\end{minipage}%
}%
\subfigure[SSRN-TPPI]{
\begin{minipage}[t]{0.15\linewidth}
\centering
\includegraphics[width=0.8\linewidth]{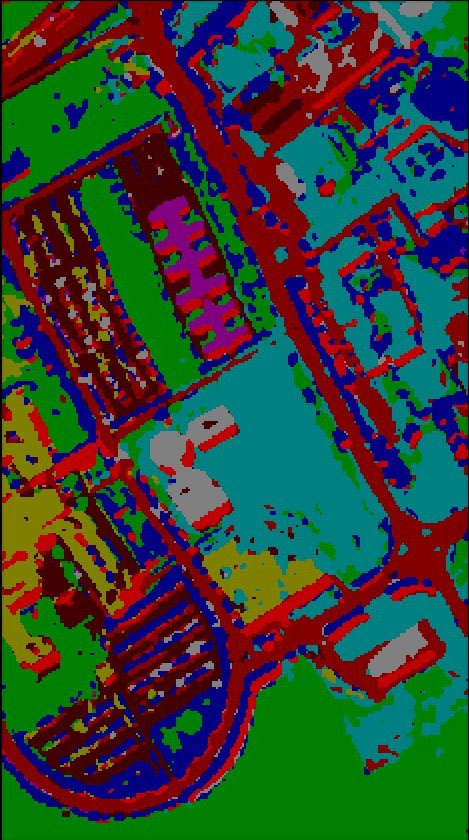}
\end{minipage}%
}%
\subfigure[pResNet]{
\begin{minipage}[t]{0.15\linewidth}
\centering
\includegraphics[width=0.8\linewidth]{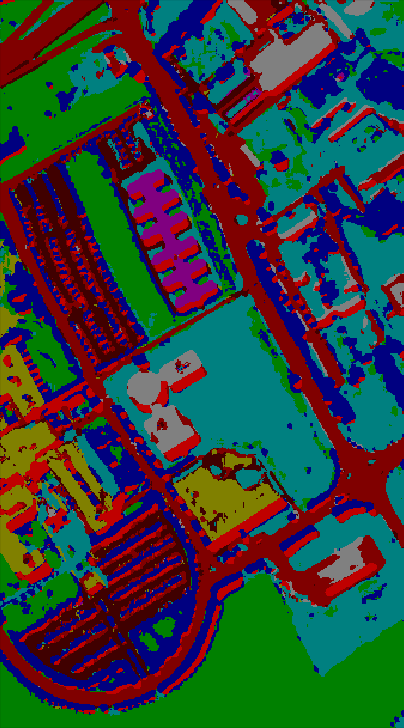}
\end{minipage}%
}%
\subfigure[pResNet-TPPI]{
\begin{minipage}[t]{0.15\linewidth}
\centering
\includegraphics[width=0.8\linewidth]{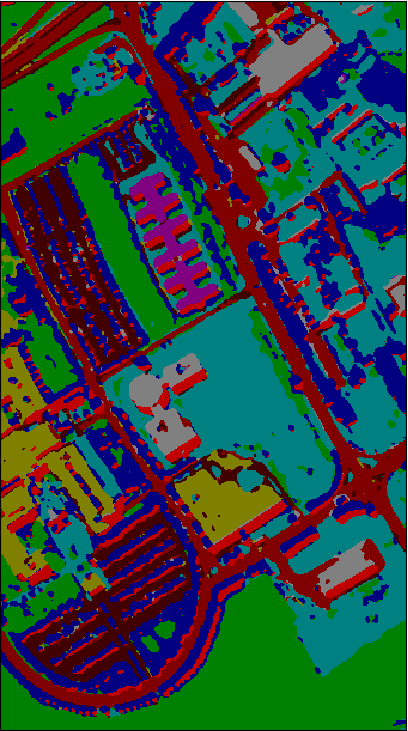}
\end{minipage}%
}%
\subfigure[GT]{
\begin{minipage}[t]{0.15\linewidth}
\centering
\includegraphics[width=0.8\linewidth]{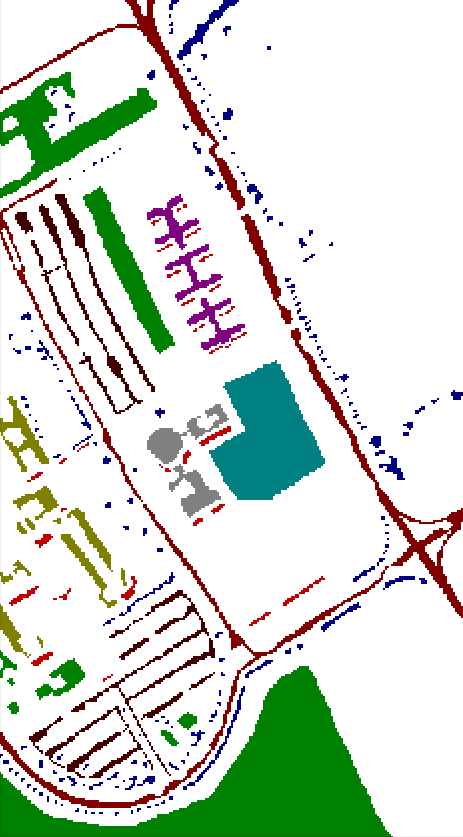}
\end{minipage}%
}%

\subfigure[SSRN]{
\begin{minipage}[t]{0.15\linewidth}
\centering
\includegraphics[width=0.8\linewidth]{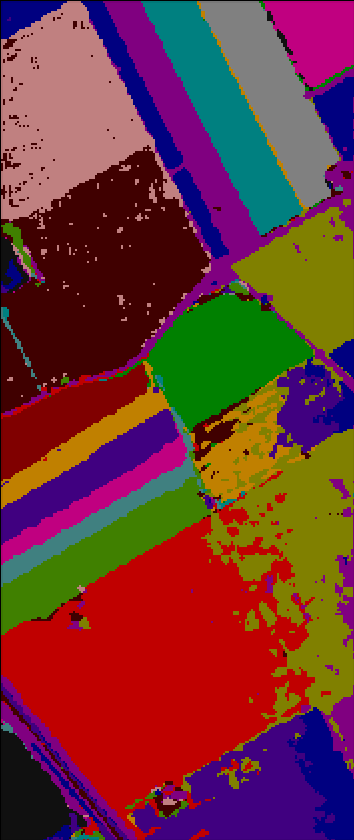}
\end{minipage}%
}%
\subfigure[SSRN-TPPI]{
\begin{minipage}[t]{0.15\linewidth}
\centering
\includegraphics[width=0.8\linewidth]{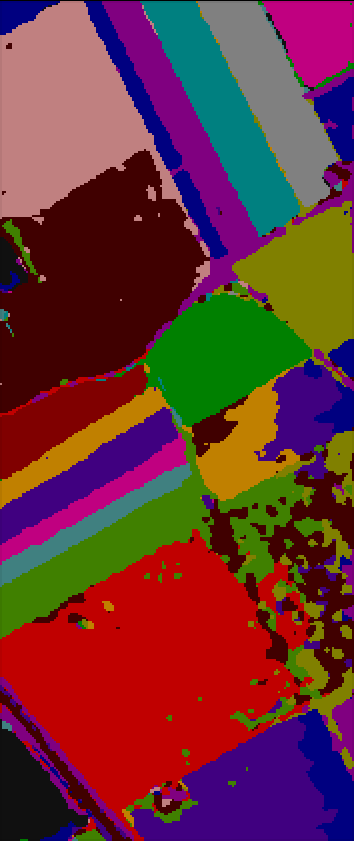}
\end{minipage}%
}%
\subfigure[pResNet]{
\begin{minipage}[t]{0.15\linewidth}
\centering
\includegraphics[width=0.8\linewidth]{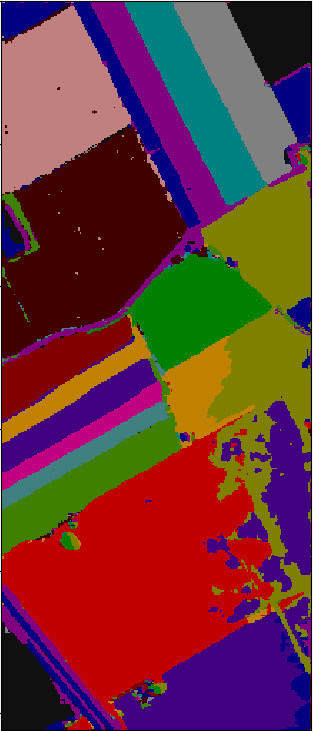}
\end{minipage}%
}%
\subfigure[pResNet-TPPI]{
\begin{minipage}[t]{0.15\linewidth}
\centering
\includegraphics[width=0.8\linewidth]{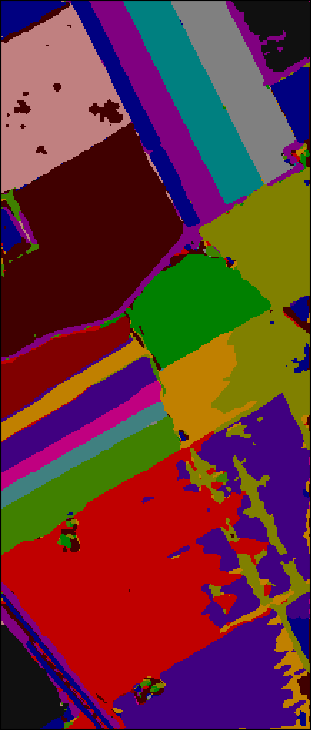}
\end{minipage}%
}%
\subfigure[GT]{
\begin{minipage}[t]{0.15\linewidth}
\centering
\includegraphics[width=0.8\linewidth]{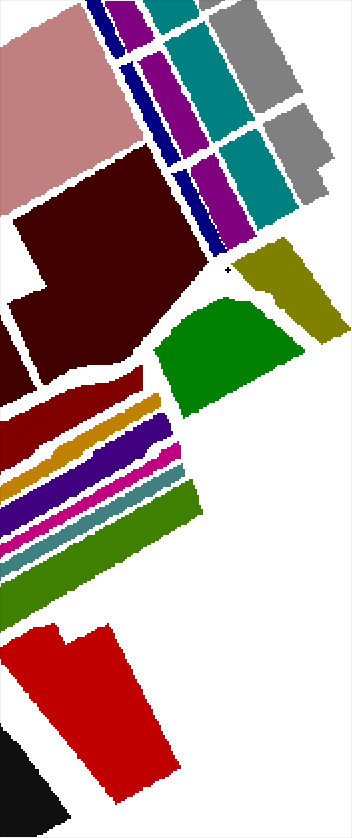}
\end{minipage}%
}%
\centering
\caption{Classification maps by SSRN and SSRN-TPPI, pResNet and pResNet-TPPI and GT (Ground Truth) on different dataset.IP:(a)-(e) PU:(f)-(j) SV:(k)-(o).}
\end{figure*}

\textbf{Classification accuracy} : We use $7*7$ spatial neighbors of pixel as the input sample to train networks. Table1,2,3 report the classification results on IP, PU, and SV dataset in test and prediction phase. 

It can be seen that: on IP, PU and SV dataset, the classification accuracy of the TPPI-Net (SSRN-TPPI and pResNet-TPPI) is almost the same as that of the corresponding TPPP-Net (SSRN and pResNet) both in the test and prediction phase.

RGB image dense prediction calls for multiscale contextual reasoning in combination with full resolution output. The reason for such conflicting demands of multi-scale reasoning and full-resolution dense prediction, we think, is because the semantic definition in the natural image semantic segmentation orient to natural objects, such as people, bicycles, houses, animals, etc. So, the discriminable features are mainly concerned with contours and shapes which usually call for multiscale contextual reasoning. However, the semantic definition in the HSI  mainly aim to land covers, such as wheat fields, woods, rivers, etc. The discriminable features are mainly concerned with texture, that is, repeated patterns. These texture features can be extracted from smaller patches at the original scale. Therefore, the accuracy will not be affected even if we have replaced the downsampling layer in pResNet.

\begin{table*}[t]
  \centering
  \caption{Predicting time (s) on the IP, PU, and SV dataset using SSRN and SSRN-TPPI.}
  \resizebox{0.8\linewidth}{10mm}{
    \begin{tabular}{p{4.19em}p{5.5em}cp{5.5em}cp{4.19em}cp{4.19em}c}
    \toprule
    \toprule
    \multicolumn{1}{r}{} & \multicolumn{2}{p{11em}}{SSRN} & \multicolumn{2}{p{11em}}{SSRN-TPPI} & \multicolumn{2}{p{8.38em}}{SSRN} & \multicolumn{2}{p{8.38em}}{SSRN-TPPI} \\
    \multicolumn{1}{r}{} & \multicolumn{2}{p{11em}}{(CPU)} & \multicolumn{2}{p{11em}}{(CPU)} & \multicolumn{2}{p{8.38em}}{(GPU)} & \multicolumn{2}{p{8.38em}}{(GPU)} \\
    \midrule
    IP    & \multicolumn{2}{p{11em}}{152.078±3.659} & \multicolumn{2}{p{11em}}{2.722±0.116} & \multicolumn{2}{p{8.38em}}{7.091±0.044} & \multicolumn{2}{p{8.38em}}{1.173±0.041} \\
    PU    & \multicolumn{2}{p{11em}}{863.114±31.363} & \multicolumn{2}{p{11em}}{15.108±0.887} & \multicolumn{2}{p{8.38em}}{49.207±2.574} & \multicolumn{2}{p{8.38em}}{1.271±0.081} \\
    SV    & \multicolumn{2}{p{11em}}{920.783±35.503} & \multicolumn{2}{p{11em}}{16.540±0.633} & \multicolumn{2}{p{8.38em}}{37.291±1.260} & \multicolumn{2}{p{8.38em}}{1.317±0.042} \\
    \bottomrule
    \bottomrule
    \end{tabular}}%
  \label{tab:addlabel}%
\end{table*}%

\begin{table*}[t]
  \centering
  \caption{Predicting time (s) on the IP, PU, and SV dataset using pResNet and pResNet-TPPI.}
  \resizebox{0.8\linewidth}{10mm}{
    \begin{tabular}{p{4.19em}p{5.5em}cp{5.5em}cp{4.19em}cp{4.19em}c}
    \toprule
    \toprule
    \multicolumn{1}{r}{} & \multicolumn{2}{p{11em}}{pResNet} & \multicolumn{2}{p{11em}}{pResNet-TPPI} & \multicolumn{2}{p{8.38em}}{pResNet} & \multicolumn{2}{p{8.38em}}{pResNet-TPPI} \\
    \multicolumn{1}{r}{} & \multicolumn{2}{p{11em}}{(CPU)} & \multicolumn{2}{p{11em}}{(CPU)} & \multicolumn{2}{p{8.38em}}{(GPU)} & \multicolumn{2}{p{8.38em}}{(GPU)} \\
    \midrule
    IP    & \multicolumn{2}{p{11em}}{29.924±1.038} & \multicolumn{2}{p{11em}}{1.292±0.055} & \multicolumn{2}{p{8.38em}}{5.487±0.440} & \multicolumn{2}{p{8.38em}}{0.804±0.025} \\
    PU    & \multicolumn{2}{p{11em}}{231.300±7.269} & \multicolumn{2}{p{11em}}{13.246±0.759} & \multicolumn{2}{p{8.38em}}{47.810±3.574} & \multicolumn{2}{p{8.38em}}{1.181±0.056} \\
    SV    & \multicolumn{2}{p{11em}}{157.392±5.922} & \multicolumn{2}{p{11em}}{6.00±0.198} & \multicolumn{2}{p{8.38em}}{30.836±1.577} & \multicolumn{2}{p{8.38em}}{1.173±0.104} \\
    \bottomrule
    \bottomrule
    \end{tabular}}%
  \label{tab:addlabel}%
\end{table*}%

According to the experimental results, we can conclude that the classification accuracy of the TPPI network is not inferior to that of the current state of the art HSI classification network.

Figure 6 shows the classification maps by using SSRN and SSRN-TPPI, pResNet and pResNet-TPPI on the IP, PU and SV dataset. As it can be observed, the proposed method provides spatially consistent classification outputs with well-delineated object borders and very few classification interferers.

\textbf{Prediction speed} : We use $7*7$ spatial neighbors of pixel as the input sample to train networks. Table 4,5 provide the predicting time in seconds for the real HSI on IP, PU, and SV dataset. It is worth mentioning, for SSRN and pResNet, batch processing is adopted whether it is on the GPU or the CPU to get the best prediction time.

From Table 4 we can note that the proposed SSRN-TPPI was far faster than SSRN in predicting speed, and the same conclusion can be deduced for pResNet and pResNet-TPPI from Table 5. 

As a result, on the IP dataset, SSRN-TPPI is approximately fifty five times and six times faster than SSRN on CPU and GPU respectively. pResNet-TPPI is approximately twenty three times and six times faster than pResNet on CPU and GPU respectively. 

On the PU dataset, SSRN-TPPI is approximately fifty seven times and thirty eight times faster than SSRN on CPU and GPU respectively. pResNet-TPPI is approximately seventeen times and forty times faster than pResNet on CPU and GPU respectively.

On the SV dataset, SSRN-TPPI is approximately fifty five times and twenty-eight times faster than SSRN on CPU and GPU respectively. pResNet-TPPI is both approximately twenty six times faster than pResNet on CPU and GPU.

The results in the table are consistent with the previous theoretical analysis, although they are not identical. The reason, we think, is because the theoretical analysis only considers the floating-point addition and multiplication in the network, and does not consider other time consumption such as memory access.

\textbf{Effect of patch size on the  classification accuracy and prediction time} : Figure 7 illustrates the change of classification accuracy and prediction time with the growth of the training patch size on IP dataset. From Figure 7(a), it is shown that the classification accuracy of both SSRN-TPPI and SSRN increases with the growth of the training patch size, but when the patch size greater than 7, the growth of OA slows down. This further proved that the discriminable features for HSI classification can be extracted in a smaller receptive field. From Figure 7(b), one can note, as the patch size increasing, the prediction time of SSRN-TPPI remain almost unchanged, but the prediction time of SSRN increases rapidly. It is reasonable since the Flops and memory access for processing a pixel will increase with the increase of patch size.
\begin{figure}[h]
\centering
\subfigure[]{
\begin{minipage}[t]{0.5\linewidth}
\centering
\includegraphics[width=0.7\linewidth]{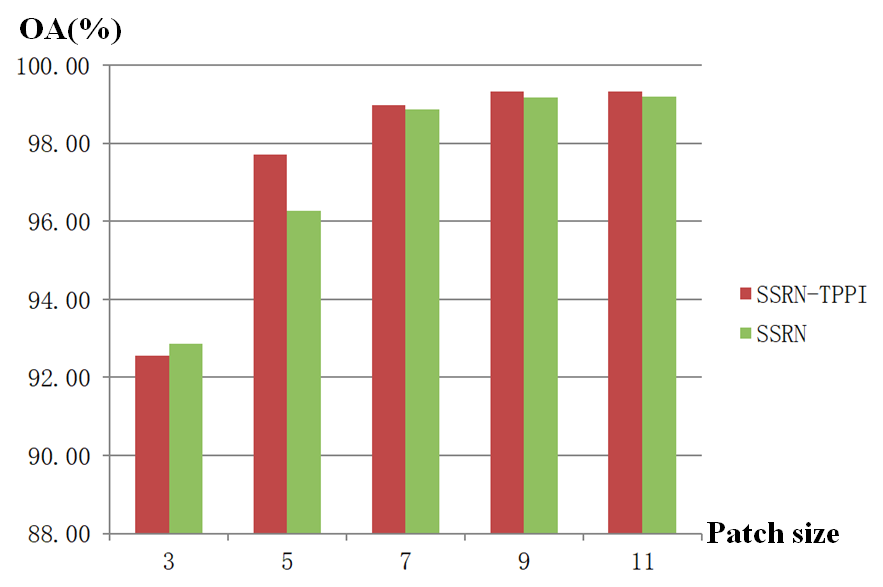}
\end{minipage}%
}%
\subfigure[]{
\begin{minipage}[t]{0.5\linewidth}
\centering
\includegraphics[width=0.8\linewidth]{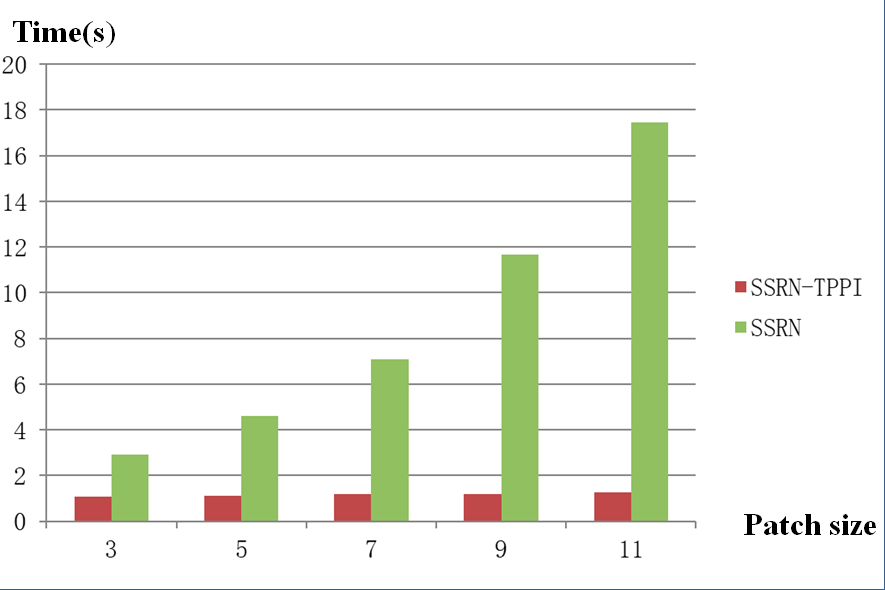}
\end{minipage}%
}%
\centering
\caption{The change of classification accuracy(a) and prediction time (b) on the IP dataset using SSRN and SSRN-TPPI with different patch size.}
\end{figure}

\section{Conclusion}

In this work, we overcome the limitations of HSI classification dataset on traditional semantic segmentation networks, and propose a brand new design mechanism TPPI suitable for efficient and practical HSI classification which can take into account both network training and actual image prediction. Experiments show that, compared with the state or the art spectral-spatial based DCNN, the prediction speed of TPPI-Net is much faster than that of corresponding DCNN while the prediction accuracy is almost unchanged.

{\small
\bibliographystyle{ieee_fullname}
\bibliography{TPPI}
}

\end{document}